\documentclass[10pt]{amsart}
\usepackage[usenames,dvipsnames]{xcolor}
\usepackage{amsmath,amsthm,amssymb,color,comment,csquotes,enumerate,fancyhdr,filecontents,graphicx,stmaryrd,verbatim}
\usepackage[round]{natbib}

\usepackage{hyperref}\hypersetup{colorlinks=true,linkcolor=MidnightBlue,citecolor=MidnightBlue,urlcolor=MidnightBlue}
\newtheorem{thrm}{Theorem}
\theoremstyle{definition}

\newtheorem{defn}[thrm]{Definition}

\DeclareMathOperator*{\argmin}{arg\,min}

\DeclareRobustCommand{\VAN}[3]{#2} % proper Dutch 'van/de' capitalisation

\begin{document}

\title[Statistical learning theory and Occam's razor]{Statistical learning theory and Occam's razor: \\ The core argument}
\author[Sterkenburg]{Tom F.\ Sterkenburg}
\date{November 21, 2024. This is the final version, as accepted for publication in \emph{Minds and Machines} (\href{https://dx.doi.org/10.1007/s11023-024-09703-y}{doi:10.1007/s11023-024-09703-y}). This research was supported by the German Research Foundation (DFG)---Projektnummern 437206810, \emph{Die Epistemologie der Statistischen Lerntheorie}; 511917847, \emph{Vom Bias zum Wissen: Die Epistemologie des maschinellen Lernens}. For helpful discussion and feedback I  thank Jan-Willem Romeijn, Daniel Herrmann, Oliver Buchholz, Konstantin Genin, Gordon Belot, and two anonymous reviewers.}
\address{Munich Center for Mathematical Philosophy, LMU Munich}
\email{tom.sterkenburg@lmu.de}

\begin{abstract}
Statistical learning theory is often associated with the principle of Occam's razor, which recommends a simplicity preference in inductive inference. This paper distills the core argument for simplicity obtainable from statistical learning theory, built on the theory's central learning guarantee for the method of empirical risk minimization. This core ``means-ends'' argument is that a simpler hypothesis class or inductive model is better because it has better learning guarantees; however, these guarantees are model-relative and so the theoretical push towards simplicity is checked by our prior knowledge.

\end{abstract}

\maketitle

\textsuperscript{}

\section{Introduction}

Statistical learning theory is the standard framework for the mathematical analysis of machine learning methods \citep{Vap00,ShaBen14}. The framework offers theoretical learning guarantees for certain learning methods, thus providing a basis for viewing such methods as \emph{good} methods.

An old trope in machine learning, usually evoked under the label of \emph{Occam's razor}, is that a shared trait of good methods is a bias towards \emph{simplicity} (\citealp{Mit97,DudHarSto01,ShaBen14,GooBenCou16,MohRosTal18,Alp20}). Occam's razor, understood as the principle that a simplicity preference is integral to good scientific or inductive reasoning, is also a long-standing topic of debate in the philosophy of science \citep{Sob15,Bak22sep}. The central question here is whether we actually have some epistemic \emph{justification} for Occam's razor. That is, we seek a rational reason for holding that a simplicity preference helps us attain desirable epistemic ends, like minimizing error.
 
A step forwards in the wider debate would be a justification for Occam's razor in machine learning methods, and an obvious place to look for such a justification is statistical learning theory. Indeed, formal results in this framework have been quoted in support of such a justification (e.g., \citealp{BEHW87ipl,ShaBen14}). On the other hand, computer scientists and philosophers alike have also relied on formal results to argue \emph{against} such a justification in this framework (e.g., \citealp{Dom99dmkd,Her20pos}). %This debate has run into something of a stalemate. Both sides rely on prima facie plausible intuitions, that indeed find support in precise mathematical expression. Yet it can be seen as a weakness in the arguments from both sides that they fail to account for the intuitions from the other side.

In this paper, I integrate the intuitions and arguments from both sides into a qualified epistemic justification for Occam's razor from statistical learning theory. Importantly, the notion of simplicity in this justification pertains, \emph{not} to individual classifiers or hypotheses, but to inductive models or \emph{classes} of hypotheses. A further important component of my account is the relativity to such inductive models of the justification obtainable from theoretical learning guarantees, as highlighted by \citet{SteGru21syn}. This \emph{model-relativity} of learning-theoretic justification aligns well, I argue, with a broad tradition in the philosophy of science which accepts the impossibility of absolute justification, and shifts attention to the project of how to rationally proceed from our current beliefs and assumptions. 
A final important characteristic of my account is the \emph{means-ends} nature of the justificatory reasoning. In one line, the means-ends justificatory argument says that in order to have better model-relative learning guarantees, we need to codify our assumptions in the form of a simpler inductive model.
%The result---with a label that is testimony to its highly qualified nature---is a \emph{means-ends model-relative justification} for Occam's razor in statistical learning theory.

This argument is based on the first of the two ``inductive principles'' \citep{Vap00} central to statistical learning theory, namely the method of \emph{empirical risk minimization}. I think this is the ``core argument''  for simplicity in statistical learning theory, which further underpins  the method of \emph{structural risk minimization} and its characteristic simplicity preference. I sketch this at the end of the paper. %The latter method may indeed seem a more natural  candidate to base an account of Occam's razor on, if only for its kinship to methods in model selection that are central to the modern philosophical literature on the topic \citep{Sob15,Rom17pos}. 
%This paper's argument based on empirical risk minimization already forms the core, I think, for an generalized justificatory argument based on structural risk minimization. 
%But that is work for another paper. 

The plan of the paper is as follows. In section \ref{sec:formals}, I present the framework of statistical learning theory and the main technical ingredients for the core argument for Occam's razor. These include the notions of empirical risk minimization, learnability, uniform convergence, and VC dimension, and the fundamental theorem that ties these notions together. In section \ref{sec:vcdimsimpl}, I argue that VC dimension is a robust notion of the simplicity of a hypothesis class. In section \ref{sec:justsimpl}, I discuss the theoretical justification for empirical risk minimization and particularly its model-relative nature; and  I show how all of the previous comes together into a justificatory argument for simplicity. %This argument relies, as mentioned, on the model-relativity of learning-theoretic justification, and on an epistemological perspective on machine learning methods that calls for this type of normative element. 
I conclude in section \ref{sec:concl}.

\subsubsection*{Motivation}
Before starting, there is a worry about the paper's general project that I should acknowledge. This worry is that the project engages with a debate of a bygone era. It has been well over a decade since \citet{HarKul07} initiated a  small wave of philosophical interest in statistical learning theory, and \citet[p.\ 860]{Ste11incl} concluded that the theory is ``worthy of further sustained interest from philosophers of science.'' This sustained interest has not exactly materialized, while the landscape of machine learning has altered significantly. Especially the advent of deep neural networks (DNN's) has caused a shift in what seem the more pertinent epistemological issues: from the traditional questions around the reliability of inductive inference to questions around interpretability and explainability \citep{BeiRaz22pc}. %,FreKonxxxai}. 
Moreover, the advent of these algorithms has problematized the very utility of %theoretical analysis \citep{ScuSnoeWilRah18iclr}, and the utility of 
statistical learning theory. % in particular. 
The theory seems simply not equipped to explain the generalization behaviour of learning methods like DNN's \citep{Bel21ac,HarRec22, BGKP22inc}, prompting a ``quest for a new framework for a `theory of induction'{''} \citep[p.\ 217]{Bel21ac}. Putting it bluntly: why a renewed philosophical engagement with the framework of statistical learning theory, if this framework is starting to look like a thing of the past?

One plain answer is that it is still of interest if and how the standard framework already offers justification for Occam's razor. % the problem of Occam's razor in the standard framework is still wide open. 
Curiously, statistical learning theory is largely left out the modern shift of the  philosophical debate towards various frameworks in mathematical statistics (\citealp[ch.\ 2]{Sob15}; \citealp[sect.\ 5]{Bak22sep}), with Sober \citeyearpar[p.\ 140, fn.\ 61]{Sob15} perceiving statistical learning theory to be ``dramatically'' different from the ``Bayesian and frequentist ideas'' that have informed this debate so far.  The current project thus fills a gap in the philosophical literature. Secondly, I may above have put things overly bluntly: it is not at all clear that core components of statistical learning theory will not continue to play an essential role in newer theory (cf.\ \citealp{BarMonRak21ac}). 
%
% .\footnote{Statistical 
%	learning theory is perceived by Sober \citeyearpar[p.\ 140]{Sob15} to be ``dramatically'' different from the ``Bayesian and frequentist ideas'' that have informed this modern debate so far.} 
%	 The current project thus fills a clear gap in the philosophical literature. 
%that these developments do not detract from the lingering question whether the standard framework already offers justification for Occam's razor, a question of enduring interest in light of the modern shift of the  philosophical debate towards various frameworks in mathematical statistics (\citealp[ch.\ 2]{Sob15}; \citealp[sect.\ 5]{Bak22sep}).\footnote{Moreover,
	%the framework of statistical learning theory is perceived by Sober \citeyearpar[p.\ 140]{Sob15} to be ``dramatically'' different from the ``Bayesian and frequentist ideas'' that have informed this modern debate so far.} 
	In any case, finally, the current project is a stepping stone towards the philosophical analysis of any new ``framework for a theory of induction'' in machine learning. Belkin indeed evokes ``a very pure form of Occam's razor'' as the ``guiding principle'' in a new framework \citeyearpar[p.\ 218]{Bel21ac}. To assess the role and justification of simplicity in such an emerging new framework, it will at the very least be helpful to actually have clarity on its role in the standard framework. The  ``generalization puzzle'' \citep[p.\ 25]{BGKP22inc} that is now hotly debated in the machine learning community is indeed a modern reincarnation of exactly those traditional philosophical questions around the reliability of induction. Work like the current project can, I hope, offer a starting point for philosophers to engage with this  exciting but complex debate.

%However, this ``generalization puzzle'' touches again exactly on the traditional philosophical questions of the reliability of induction, and for philosophers to assess and perhaps even take part in these new developments, it seems at the very least helpful to first attain conceptual clarity on the traditional framework. This holds, in particular, for the role of Occam's razor: in fact, Belkin himself evokes ``a very pure form of Occam's razor'' as the ``guiding principle'' in a new framework \citeyearpar[p.\ 218]{Bel21ac}. I also see the current paper as a stepping stone towards a philosophical analysis of any newly emerging ``framework for a theory of induction.'' 

%the traditional philosophical questions around the reliability of inductive inference are exactly what is at stake in the debate around the generalization of DNN's, and philosophical engagement with 

%theoretical guarantees
%
%question of justification through theory: guarantee of SRM with simplicity preference a justification for Occam's razor? [anywhere stated like this? ShaBen14?]
%
%[is this also the form of the 1987 Occam's razor argument?]

%\section{The argument from empirical risk minimization}

\section{The formal ingredients}\label{sec:formals}
In my presentation of the framework of statistical learning theory, I mainly follow \citet{ShaBen14}.\footnote{Their
	presentation is essentially a synthesis of Vapnik's \citeyearpar{Vap99ieeenn,Vap98,Vap00} ``general setting of learning'' and Valiant's \citeyearpar{Val84acm} model of ``probably approximately correct'' (PAC) learning \citep[p.\ 28]{ShaBen14}. The main concern in Vapnik's setting is the statistical analysis of uniform convergence of learning algorithms, and this approach is also simply called \emph{VC theory} after the groundbreaking early work of \citet{VapChe71tpa}. 	The tradition initiated by Valiant is also called \emph{computational learning theory} (see \citealp{AntBig92,KeaVaz94}), and an essential component is the computational efficiency of learning. This computational component is separated from the statistical component in Shalev-Shwartz and Ben-David's presentation, and I will likewise not be concerned with computational considerations in this paper.}\textsuperscript{,}\footnote{A chapter-length introduction to statistical learning theory aimed at philosophers, that I also draw from, is \citep{LuxSch11inc}. A more basic philosophical introduction is \citep{HarKul07}.}
	I restrict attention to the most basic learning paradigm in this framework, the paradigm of binary classification.

\subsection{Binary classification}
In this type of learning problem, we have a domain $\mathcal{X}$ of \emph{instances} (say, images of animals). We seek to assign these instances binary \emph{labels} (say,  \textsf{cat} or \textsf{not cat}). More precisely, we seek a general classifier or \emph{hypothesis} $h: \mathcal{X} \rightarrow \mathcal{Y}$ that maps all instances in $\mathcal{X}$ to a label in the binary label set $\mathcal{Y}$.\footnote{Hypotheses
	are often called \emph{models} in the machine learning literature. I will stick here to the terminology of \citet{ShaBen14}, also to not risk confusion with the notion of \emph{inductive model} (class of hypotheses) in the model-relative justification I discuss in section \ref{sec:justsimpl}.}

This is a learning problem because we first draw a finite \emph{training sample} of labeled instances, on the basis of which we then seek to find---to \emph{learn}---a general hypothesis. The assumption in statistical learning theory is that there always is some true but unknown distribution $\mathcal{D}$ over $\mathcal{X} \times \mathcal{Y}$, that governs both the sampling of instances and (via the conditional $\mathcal{D}(\mathcal{Y} \mid \mathcal{X})$)  the connection between instances and labels. It is assumed we obtain labeled instances by repeatedly drawing from this same distribution: the labeled instances are \emph{independently and identically distributed} (i.i.d.). In this way, we draw a training sample $S$, that is a finite ordered sequence of input-label pairs. %, or formally, $S \in  \cup_{m\in \mathbb{N}} (\mathcal{X} \times \mathcal{Y})^m$. 
Based on the training sample, we seek to learn a good hypothesis.

To assess hypotheses, we use some error function. The standard choice in binary classification is the 0/1 error function, that returns error 0 (error 1) for a correct (incorrect) classification.  Then the \emph{empirical error} of $h$ on a sample $S$ is given by the mean 0/1 error of instances,
\begin{align}
L_S(h) := \frac{|\{(x,y)\in S: h(x) \neq y \}|}{|S|}.
\end{align}
But what we are actually interested in is the quality of a classifier over all possible instances. We express this as the expected error or  \emph{risk} of $h$ with respect to the true distribution $\mathcal{D}$ over $\mathcal{X} \times \mathcal{Y}$,  
\begin{align}
L_\mathcal{D}(h) := \mathbb{E}_{(X,Y) \sim \mathcal{D}}\left[L_{(X,Y)}(h) \right].
\end{align}
We thus seek to find a hypothesis, based on a training sample $S$, with a low risk with respect to the true but unknown distribution $\mathcal{D}$.

\subsection{Hypothesis classes and learning methods}
In the framework of statistical learning theory, we are fully agnostic about the shape of the distribution $\mathcal{D}$. However (as I discuss in more detail later), we cannot get anywhere unless we impose restrictions elsewhere. The approach in statistical learning theory is to make the analysis relative to some hypothesis class $\mathcal{H}$. We then seek to select a hypothesis $h$ from $\mathcal{H}$ which has \emph{relatively} low risk, among those hypotheses in $\mathcal{H}$, with respect to true but unknown $\mathcal{D}$. That is, we seek to select, based on training data, a hypothesis with risk close to $\min_{h' \in \mathcal{H}}L_\mathcal{D}(h')$.\footnote{This	
	is what \citet[p.\ 23]{ShaBen14} call \emph{agnostic} learning, as opposed to the more specific paradigm of \emph{realizable} learning, where we make the (very strong) assumption that $\mathcal{H}$ already contains an $h^*$ with zero true risk (ibid., def.\ 2.3). In computational learning theory this assumption is actually standard.}

%\subsubsection{Learning methods}
This is a machine learning problem because we want to specify an automated learning procedure or \emph{learning method} to do this selection---this \emph{learning}---of hypotheses from samples. Formally, we treat a learning method as a function from all possible samples to hypotheses.\footnote{Again, 
	I abstract away from computational considerations. A restriction of the framework to formal computability (in which learning methods are actual algorithms, i.e., Turing-computable functions; a framework only first studied recently, \citealp{aablu20alt}) does not appear to substantially change the notions and results discussed here \citep{Ste22colt}.}

A basic such method is the procedure of \emph{empirical risk minimization} (\textsc{ERM}) for given hypothesis class $\mathcal{H}$. This method simply selects for given sample $S$ a hypothesis in $\mathcal{H}$ with minimal error on the sample.
\begin{defn}
Empirical risk minimization for hypothesis class $\mathcal{H}$, write \textsc{ERM}$_\mathcal{H}$, returns for each $S \in \mathcal{S}$ a hypothesis in $\argmin_{h \in \mathcal{H}}L_S(h)$.
\end{defn}

What makes a learning method like \textsc{ERM} for $\mathcal{H}$ a \emph{good} method? Given the indicated goal of finding a relatively-low-risk hypothesis in $\mathcal{H}$, method \textsc{ERM}$_\mathcal{H}$ can be called good if it has some sufficiently strong guarantee of attaining this goal. %; or rather, of attaining some formally precise version of this informal goal. %It is customary to say that a method that provably meets some such precise guarantee \emph{learns} the hypothesis class in the corresponding precise sense, and that the hypothesis class is therefore \emph{learnable} in that same sense.

%answer this question, we need some (feasible!) criterion of successful learning. 

\subsection{Learnability} \label{sssec:learnty}
The main formal guarantee of good learning is formulated in terms of the following components.

First, we quantify the ``relatively-low-risk'' by an accuracy parameter $\epsilon$. This $\epsilon$ bounds the difference between the best possible risk $\min_{h \in \mathcal{H}} L_\mathcal{D}(h)$ and the risk $L_\mathcal{D}(A_\mathcal{H}(S))$ of a hypothesis selected by method $A_\mathcal{H}$ on sample $S$. This difference is also called the \emph{estimation error}.

Second, because of the randomness in the generation of samples from $\mathcal{D}$, any guarantee can at best be probabilistic. Intuitively, we can only expect a learning method to select a good hypothesis based on samples that are in fact representative of the true distribution $\mathcal{D}$; but we cannot exclude that with small probability we draw a sample that is not representative. Hence we also introduce a confidence parameter $\delta$ that quantifies this probability. 

Finally, again due to the randomness in drawing samples, the quality of an estimate is inevitably connected to the size of the sample. We will thus formulate our guarantee as a relation between sample size, confidence, and accuracy. 

This guarantee is  \emph{probably approximately correct} (PAC) \emph{learnability}\footnote{This 
	notion was formulated (with the additional component of computational complexity) by \citet{Val84acm}, while the term ``pac-learning'' appears to be due to \citet{AngLai88ml}. %The notion (as presented here, without the computational component) can be seen as a reformulation of the notion of uniform convergence studied in VC theory (\citealp[p.\ 201]{DevGyoLug96}).
	}---or simply, \emph{learnability}. %In this definition the above components are related as follows. 
Hypothesis class $\mathcal{H}$ is \emph{learnable} by a learning method $A_\mathcal{H}$ if for any given inaccuracy $\epsilon$ and confidence $1-\delta$, there is a large enough sample size $m_0$ such that for any $m \geq m_0$, we have the following, no matter the true distribution $\mathcal{D}$. With probability at least $1-\delta$ over the possible size-$m$ samples $S^m$ drawn i.i.d.\ from $\mathcal{D}$,  method $A_\mathcal{H}$, on receiving such a sample, returns a hypothesis with estimation error below $\epsilon$. %that has risk no higher than $\epsilon$ than the lowest-risk hypothesis in $\mathcal{H}$. 
To rephrase,

\begin{defn}[Learnability]\label{def:uniflearn}
A hypothesis class $\mathcal{H}$ is learnable if there exists a learning method $A_\mathcal{H}: \mathcal{S} \rightarrow \mathcal{H}$ and a sample size function $m_\mathcal{H}: (0,1)^2 \rightarrow \mathbb{N}$  such that for all $\epsilon,\delta \in (0,1)$, for all $m \geq m_\mathcal{H}(\epsilon,\delta)$ and any distribution $\mathcal{D}$ over $\mathcal{X} \times \mathcal{Y}$,
\begin{align}\label{eq:pac}
\textrm{Prob}_{S \sim \mathcal{D}^m}\left[L_\mathcal{D}(A_\mathcal{H}(S)) \leq \min_{h \in \mathcal{H}} (L_\mathcal{D}(h))+ \epsilon \right] \geq 1-\delta.
\end{align}
\end{defn}

Note that this guarantee (in particular, the minimum sample size $m_\mathcal{H}(\epsilon,\delta)$ for given $\epsilon$ and $\delta$) only depends on the hypothesis class $\mathcal{H}$. In line with the agnostic approach of statistical learning theory, it is a \emph{distribution-free} guarantee: the sample size does not depend  on the true distribution $\mathcal{D}$. 
%Nor does the sample size depend on what is actually the best hypothesis in $\mathcal{H}$. The $\epsilon$-bound holds for \emph{all} $h$ in the hypothesis class: in definition \ref{def:uniflearn}, instead of \eqref{eq:pac}, I could also have written  
%\begin{align}\label{eq:pac2}
%\textrm{Prob}_{S \sim \mathcal{D}^m}\left[(\forall h \in \mathcal{H}) \left[ L_\mathcal{D}(A_\mathcal{H}(S)) \leq  L_\mathcal{D}(h)+ \epsilon  \right] \right] \geq 1-\delta.
%\end{align}
%
%We say that the sample size in this guarantee is \emph{uniform} with respect to the hypotheses in $\mathcal{H}$, and since this distinguishes the current notion of learnability from the one I discuss in sect.\ \ref{sec:nonunif}, I call this guarantee uniform learnability.%\footnote{uniform convergence central to VC theory.}

Learnability is much related to another property of a hypothesis class, namely \emph{uniform convergence}. The former, as we have seen, concerns the estimation error of a learning method; the latter concerns the difference between the empirical errors and the true risks of all hypotheses in the class. This property will allow us to relate the \textsc{ERM} method to learnability.

\subsection{Uniform convergence and empirical risk minimization}\label{sssec:unifconv}
The law of large numbers already tells us that, for any fixed hypothesis $h$, as we draw larger and larger samples $S^m$ i.i.d.\ from true distribution $\mathcal{D}$, the empirical error of $h$ on $S^m$ will in probability converge to its true risk. However, in our learning problem, we are not interested in fixing a particular hypothesis and estimating its true risk. We are interested in the performance of a learning algorithm, which, depending on the data, can select different hypotheses. For this we need something stronger, namely a ``uniform law of large numbers,'' which bounds the difference between empirical errors and true risks of all hypotheses \emph{uniformly}---simultaneously. 

For given hypothesis class $\mathcal{H}$, call a training sample $\epsilon$-\emph{representative}  if simultaneously for all hypotheses $h \in \mathcal{H}$ the difference between $h$'s empirical error $L_S(h)$ on $S$ and $h$'s true risk $L_\mathcal{D}(h)$ is smaller than $\epsilon$,
\begin{align}
(\forall h \in \mathcal{H}) \left[  | L_S(h) - L_\mathcal{D}(h)| \leq \epsilon \right].
\end{align}
 On such a sample, all empirical errors give good indications of the true errors: ``what you see is what you get'' (``wysiwyg'', terminology from \citealp{Bel21ac}). Now a hypothesis class has the \emph{uniform convergence property} if there is a ``wysiwyg'' guarantee of drawing such representative samples. Precisely,

%A training sample $S$ is called $\epsilon$-\emph{representable} with respect to class $\mathcal{H}$ if simultaneously for all hypotheses $h \in \mathcal{H}$ the difference between $h$'s empirical error $L_S(h)$ on $S$ and $h$'s true risk $L_\mathcal{D}(h)$ is smaller than $\epsilon$. In words, on such a sample, all empirical errors give good indications of the true errors.
%
%Now a hypothesis class has the \emph{uniform convergence property} if for any given accuracy $\epsilon$ and confidence $1-\delta$, there is a large enough sample size $m_0$ such that for any $m \geq m_0$, we have the following, no matter the true distribution $\mathcal{D}$. With probability at least $1-\delta$ over the possible size-$m$ samples $S^m$ drawn i.i.d.\ from $\mathcal{D}$, the drawn sample is $\epsilon$-representative. To rephrase,

\begin{defn}[Uniform convergence]\label{def:uniconv}
A hypothesis class $\mathcal{H}$ has the uniform convergence property if there exists a sample size function $m_\mathcal{H}^\mathrm{uc}: (0,1)^2 \rightarrow \mathbb{N}$  such that for all $\epsilon,\delta \in (0,1)$, for all $m \geq m_\mathcal{H}^\mathrm{uc}(\epsilon,\delta)$ and any distribution $\mathcal{D}$ over $\mathcal{X} \times \mathcal{Y}$ we have
\begin{align}\label{eq:unifconv}
\textrm{Prob}_{S \sim \mathcal{D}^m}\left[(\forall h \in \mathcal{H}) \left[  | L_S(h) - L_\mathcal{D}(h)| \leq \epsilon \right] \right] \geq 1-\delta.
\end{align}
\end{defn}

To link this property to the \textsc{ERM} method, we first reformulate it. Namely, we have stated it in terms of the minimum sample size $m_\mathcal{H}^\textrm{uc}(\epsilon,\delta)$ we need for given $\epsilon$ and $\delta$; but  we can also formulate it as the bound on $\epsilon$ we get for given $\delta$ and sample size $m$. Precisely, there is an accuracy function $\epsilon_\mathcal{H}^\mathrm{uc}(m,\epsilon)$ such that  for given $\delta$ and $m$ we have with probability at least $1-\delta$ that 
\begin{align}
(\forall h \in \mathcal{H}) \left[ | L_\mathcal{D}(h) - L_S(h) | \leq \epsilon_\mathcal{H}^\mathrm{uc}(m,\delta) \right],
\end{align}
which in particular gives a uniform upper bound on true risk in terms of empirical error,
\begin{align}
(\forall h \in \mathcal{H}) \left[ L_\mathcal{D}(h) \leq L_S(h) + \epsilon_\mathcal{H}^\mathrm{uc}(m,\delta) \right].
\end{align}

Now recall that the method \textsc{ERM}$_\mathcal{H}$ for given sample $S$ of length $m$ selects an $h$ that minimizes $L_S(h)$. Since $\epsilon_\mathcal{H}^\mathrm{uc}(m,\delta)$ is a constant term for fixed $m$ and $\delta$, method \textsc{ERM}$_\mathcal{H}$ can be seen to explicitly minimize this upper bound 
\begin{align}
 L_\mathcal{D}(h) \leq L_S(h) + \epsilon_\mathcal{H}^\mathrm{uc}(m,\delta)
\end{align}
on the true risk. Thus, given $\mathcal{H}$ satisfies uniform convergence, \textsc{ERM}$_\mathcal{H}$ selects a hypothesis with the sharpest uniform upper bound on its true risk.

This minimization property, under the assumption of uniform convergence, allows us to derive that \textsc{ERM}$_\mathcal{H}$ \emph{learns} $\mathcal{H}$. 
%Namely, for given $\epsilon$ and $\delta$, pick $m \geq m_\mathcal{H}^\textrm{uc}(\epsilon / 2,\delta)$, so that $\epsilon_\mathcal{H}^\textrm{uc}(m,\delta) \leq \epsilon /2$. By instantiating ERM$_\mathcal{H}$ in (6), and using $L_S(\textrm{ERM}_\mathcal{H}(S))= \min_{h \in \mathcal{H}} L_S(h)$, we have that with probability at least $1-\delta$
%\begin{align}
% L_\mathcal{D}(\textrm{ERM}_\mathcal{H}(S)) \leq \min_{h \in \mathcal{H}} L_S(h) + \epsilon / 2,
%\end{align}
%and so, since $m \geq m_\mathcal{H}^\textrm{uc}(\epsilon / 2,\delta)$, 
%\begin{align}
% L_\mathcal{D}(\textrm{ERM}_\mathcal{H}(S)) \leq \min_{h \in \mathcal{H}} L_\mathcal{D}(h) + \epsilon / 2 + \epsilon / 2 = \min_{h \in \mathcal{H}} L_\mathcal{D}(h) + \epsilon.
%\end{align}
%
%Recall that the method of ERM$_\mathcal{H}$ always selects the lowest-empirical-error hypothesis in $\mathcal{H}$. This method fullfills the guarantee of uniform learnability of $\mathcal{H}$ as soon as $\mathcal{H}$ has the uniform convergence property. 
Informally,\footnote{See \citet[sect.\ 4.1, specifically lemma 4.2]{ShaBen14} for the  (straightforward) formal derivation.} if we have a guarantee that large enough samples are probably representative (uniform convergence), then in particular the lowest-empirical-error hypotheses (selected by \textsc{ERM}$_\mathcal{H}$) probably have approximately lowest true risk, and so small estimation error (learnability).

	Uniform convergence thus gives us a sufficient condition for learnability, and learnability by \textsc{ERM}. However, this is still a rather abstract property, that does not give much intuition for what kind of hypothesis classes satisfy it. Fortunately, it turns out that there is a more concrete and intuitive property of hypothesis classes that is equivalent to learnability, and in fact already equivalent to learnability by \textsc{ERM}. This property is a criterion of the \emph{simplicity} of a hypothesis class.
	
%guarantee on (uniform):
%
%* (for given m) lowest upper bound on true error of selected h
%* (for given m) upper bound on difference between true error of selected h and true error of best h
%* (for given m) upper bound on difference between true and empirical error of selected h
%
%guarantee on (nonuniform)
%
%* (for given m) lowest ([complexity-of-]h-relative) upper bound on true error of selected h
%* (for given m) upper bound on difference between true error of selected h and true error of reference h
%* (for given m) upper ([complexity-of-]h-relative) bound on difference between true and empirical error of selected h

\subsection{The VC dimension}
Take a  finite set  $X= \{x_1,\dots,x_m\} \subset \mathcal{X}$ of unlabeled instances. There are several different ways in which we can label all instances in $X$; precisely, for binary labels, there are $2^{m}$ possible such labelings. Now take a hypothesis class $\mathcal{H}$. Each hypothesis $h$ in $\mathcal{H}$ gives some such possible labeling of the instances in $X$. If the hypotheses in $\mathcal{H}$ cover \emph{all} possible labelings, that is, for each possible labeling of $X$, there is some $h \in \mathcal{H}$ that gives exactly this labeling, then we say that $\mathcal{H}$ \emph{shatters} $X$.\footnote{\label{fn:shatter}Slightly
	more formally, define the \emph{restriction of $\mathcal{H}$ to finite set $X$} as the class $\mathcal{H}_{|X}$ of functions $f: X \rightarrow \mathcal{Y}$ such that $f(x) = h(x)$ for some $h \in \mathcal{H}$ and all $x \in \mathcal{X}$. Then $\mathcal{H}$ shatters finite $X \subset \mathcal{X}$ if the restriction of $\mathcal{H}$ to $X$ contains \emph{all} functions $f: X \rightarrow \mathcal{Y}$, that is, $|\mathcal{H}_{|X}|=2^{|X|}$.}

%The crucial notion in demarcating classes that are and that are not learnable builds on a notion that expresses that a given hypothesis class $\mathcal{H}$ covers all possible labelings of a finite set of instances. Define, for given hypothesis class $\mathcal{H}$ and finite instance set $X= \{x_1,\dots,x_m\} \subset \mathcal{X}$, the \emph{restriction of $\mathcal{H}$ to finite set $X$} as the class $\mathcal{H}_{|X}$ of functions $f: X \rightarrow \mathcal{Y}$ such that $f(x) = h(x)$ for some $h \in \mathcal{H}$ and all $x \in \mathcal{X}$.
%%defined by %$\mathcal{H}_{|X}:=\{(h(x_1),\dots,h(x_m)): h \in \mathcal{H}\}$. 
%%$\mathcal{H}_{|X}:=\{f : (\exists h \in \mathcal{H})( \forall x \in X)[ h(x)=f(x)] \}$.
%Then we say that $\mathcal{H}$ \emph{shatters} finite $X \subset \mathcal{X}$ if the restriction of $\mathcal{H}$ to $X$ contains \emph{all} functions $f: X \rightarrow \mathcal{Y}$. %, that is, $|\mathcal{H}_{|X}|=2^{|X|}$.
%That is, each possible labeling of the instances in $X$ is covered by \emph{some} hypothesis in $\mathcal{H}$.

The crucial notion in demarcating classes that are and are not learnable  relies on the ability to shatter sets of instances. Namely, the \emph{Vapnik-Chervonenkis dimension} (VC dimension, after \citealp{VapChe71tpa}) of hypothesis class $\mathcal{H}$ is defined as the largest size of a subset $X$ of instances for which $\mathcal{H}$ can do so.

\begin{defn}
The \emph{VC dimension} of hypothesis class $\mathcal{H}$ %, write VCdim$(\mathcal{H})$, 
is the maximal size of a set $X \subset \mathcal{X}$ that is shattered by $\mathcal{H}$. If $\mathcal{H}$ shatters sets of arbitarily large size, then the VC dimension of $\mathcal{H}$ is infinite. A \emph{VC class} is a class with finite VC dimension. 
\end{defn}

In machine learning terminology, VC dimension is a measure of the \emph{capacity} of a hypothesis class.\footnote{There 
	exist several generalizations of VC dimension, like the Natarajan dimension in multiclass categorization (see \citealp[ch.\ 29]{ShaBen14}), and indeed altogether different capacity notions in different paradigms, like the Littlestone dimension in realizable online learning (see ibid., sect.\ 21.1), and the parametric complexity in MDL inference (see \citealp{Gru07}). An important alternative capacity notion to VC dimension for classification is Rademacher complexity, which can yield stronger data-dependent bounds (see \citealp[sect.\ 5.7]{LuxSch11inc}, \citealp[ch.\ 26]{ShaBen14}). The notion of the capacity of a function class and its relation to generalization appears to have been introduced by \citet{Cov65ieee}.} 
	It is a measure of the extent to which a hypothesis class covers---contains hypotheses with good fit on---possible data samples. In that sense VC dimension is a notion of the ``richness'' %\citep[p.\ 24]{HarKul07} 
	or \emph{complexity} of a hypothesis class; and finiteness of VC dimension a criterion of a hypothesis class being sparse or \emph{simple}. I discuss this simplicity interpretation in more detail in section \ref{sec:vcdimsimpl} below.

\subsection{Bringing it all together}\label{sssec:fundthrm}
The central result of statistical learning theory elegantly ties together the main notions of the previous sections.

\begin{thrm}[Fundamental theorem of statistical learning theory\footnote{In 
	their pioneering work, \citet{VapChe71tpa} established the link between uniform convergence, ``consistency'' of \textsc{ERM}, and their notion of VC dimension, proving a generalization of the ``fundamental theorem of mathematical statistics,'' the Glivenko-Cantelli theorem of the uniform convergence of the empirical distribution function (see \citealp[ch.\ 12]{DevGyoLug96}). The connection between VC theory and computational learning theory (in particular, Valiant's notion of PAC learnability) was first spelled out by \citet{BEHW86stoc,BEHW89jacm}.}%\textsuperscript{,}\footnote{The most significant direction is from finite VC dimension to uniform convergence. Very roughly, the proof approach as presented by \citet{ShaBen14} is as follows. The main component is a major generalization of the proof for finite classes (that essentially rests on two basic lemma's, the union bound and Hoeffding's inequality, see ibid., sect.\ 4.2) to a bound on the difference between true and empirical risk in terms of a \emph{growth function}. This growth function tracks the ``effective size'' of the hypothesis class; for each $n$, it is the largest set $\mathcal{H}_{|X}$ for any $X$ of size $n$ (see footnote \ref{fn:shatter}; ibid, def.\ 6.9). Next, we can show that for a class of VC dimension $d$, the growth function beyond $d$ grows, not exponentially, but only polynomially (Sauer's lemma, see ibid., lmm.\ 6.10). Plugging this in into the bound, the result follows (see ibid., p.\ 51). Also see \citet[sects.\ 5.1--5.6]{LuxSch11inc}; \citet[sect.\ 12.4]{DevGyoLug96}.}
]\label{thrm:funpac}
%A hypothesis class $\mathcal{H}$ is uniformly learnable if and only if ERM$_{\mathcal{H}}$ uniformly learns $\mathcal{H}$ if and only if  if and only if $\mathcal{H}$ is a VC class.
The following are equivalent:
\begin{itemize}
\item $\mathcal{H}$ has the uniform convergence property;
\item $\mathcal{H}$ is learnable;
\item $\mathcal{H}$ is learnable by \textsc{ERM}$_\mathcal{H}$;
\item $\mathcal{H}$ is a VC class.
\end{itemize}
\end{thrm}

In particular, if, and only if, hypothesis class $\mathcal{H}$ has finite VC dimension, we have a ``wysiwyg'' guarantee of  a good indication of true risk (uniform convergence), and a method, \textsc{ERM}$_\mathcal{H}$, that is a \emph{good} method, in the sense of satisfying a guarantee of minimizing estimation error (learnability). We therefore have, for VC class $\mathcal{H}$, a certain \emph{justification} for the \textsc{ERM}$_\mathcal{H}$ method.\footnote{\label{fn:modelassess1}In
	my presentation, I also follow \citet{ShaBen14} in focusing on the epistemic end of learnability (minimizing estimation error). I only note here that another important epistemic end that is supported by the uniform convergence ``wysiwyg'' guarantees is \emph{model assessment}, where we use the training error to assess whether the model (the learned hypothesis or indeed the hypothesis class) is actually good. (For instance, in the discussion of VC theory by \citealp[sect.\ 7.9]{HasTibFri09}, the emphasis is rather on this end.)} %I will continue to build the justificatory story on the inductive end of learnability, but briefly return to this in footnote \ref{fn:modelassess2}.}

Actually, we can further fine-grain this picture within the finite VC dimension regime. Namely, the VC dimension of class $\mathcal{H}$ gives a quantitative bound on the sample size for uniform convergence and for learnability.

\begin{thrm}[Fundamental theorem, quantitative version\footnote{See 
	\citet[thrm.\ 6.8]{ShaBen14}.}\textsuperscript{,}\footnote{This is a bound on the sample size in terms of given accuracy and confidence parameter (and VC dimension); but we can also infer other bounds by making other choices in what quantities we take as given and what quantity we then solve for. For instance, we can derive that there exists constant $C$ such that for any given $m$ and $\delta$ we have an accuracy bound
\begin{align}
\epsilon^\mathrm{uc}(m,\delta) < C \sqrt{\frac{\mathrm{VCdim}(\mathcal{H})-\log \delta}{m}}.
\end{align}}]
For any VC class $\mathcal{H}$, there are constants $C_1, C_2$ such that, for any $\epsilon, \delta$, we have 
$$C_1 b \leq m_\mathcal{H}^\mathrm{uc}(\epsilon,\delta), m_\mathcal{H}(\epsilon,\delta) \leq C_2 b$$
where
$$ b = \frac{\mathrm{VCdim}(\mathcal{H})-\log \delta}{\epsilon^2}.$$
\end{thrm}

At bottom, the fundamental theorem expresses a certain relation between four quantities (VC dimension, sample size, accuracy, and confidence), where, in particular, a lower VC dimension makes room for lower values of the other three quantities (meaning, for stronger bounds). Thus, a lower VC dimension of $\mathcal{H}$ goes with a better guarantee and therefore a \emph{stronger justification} for  the \textsc{ERM}$_\mathcal{H}$ method.   

I discuss this justification, and how to further turn this into a justificatory argument for simplicity, in more detail in section \ref{sec:justsimpl} below. But first I will zoom in on the relevant notion of simplicity, given by the formal notion of VC dimension.

\section{The notion of simplicity}\label{sec:vcdimsimpl}
While, I will argue in this section, the notion of VC dimension does not give us a handle on the simplicity of  \emph{individual} hypotheses (section \ref{ssec:vchscompl}), it does constitute a plausible and robust measure of the simplicity of hypothesis classes (section \ref{ssec:vccompl}).

\subsection{Individual hypotheses and hypothesis classes}\label{ssec:vchscompl}
Capacity notions like VC dimension apply to hypothesis classes, \emph{not} individual hypotheses. Yet in some discussions of simplicity, that also rely on the relation between uniform convergence and small size or capacity of the hypothesis class,  the notion of simplicity invoked actually concerns individual hypotheses. These discussions use a notion of the complexity of an hypothesis as its representational complexity  in some formal language.

An influential example is the argument of \citet{BEHW87ipl} that ``under very general assumptions, Occam's Razor produces hypotheses that with high probability will be predictive of future observations'' (ibid., p.\ 378).\footnote{The 
	relevant result is derived within Valiant's PAC learning framework, and the requirement of computational efficiency in the definition and generalization guarantee of the relevant ``Occam-algorithm'' make it a bit more involved than the reasoning I discuss in this paper.  For further expositions, see \citet[pp.\ 59ff]{AntBig92}, \citet[ch.\ 2]{KeaVaz94}; and for a rebuttal of the argument, see \citet{Her20pos}.} 
	An earlier example still is \citet{Pea78ijgs}, who discusses the connection between simplicity and ``credibility'' of hypotheses via different notions of capacity and generalization success---including already VC dimension and uniform convergence (ibid., pp.\ 261ff). 
	
Pearl sets the stage as follows. We take some language $L$ with an interpretation function $I$ that maps sentences in the language to hypotheses (ibid., pp.\ 256f). Then we define some complexity measure on each sentence $t$, ``which may represent either the syntactic aspect of the sentence $t$, or the work required for the computation of $I(t)$'' (ibid., p.\ 257).\footnote{The 
	standard example is the two-symbol language of bits, where the complexity of a sentence (a bit string) is defined as its length.} 
	Further, ``the complexity of a [hypothesis $h$] with respect to a language $L$ is defined as the complexity of the simplest sentence which represents that [hypothesis]'' (ibid.). This allows us to take subsets of sufficiently \emph{simple} hypotheses: a ``\emph{complexity bounded sublanguage} of $L$ is a sublanguage $L_c=(T_c,I_c)$ such that $T_c \subseteq T$, $I_c \subseteq I$ and $C(h) \leq c$ for all $h \in I_c$'' (ibid., p.\ 258, slight change in notation). 

Now the lower the complexity $c$, the smaller the size (and in particular, the capacity) of the sublanguage (hypothesis class) $L_c$.\footnote{For 
	instance, for the language of bits, there can only be $2^{n+1}$ sentences (bit strings) of complexity (length) up to $n$, hence at most $2^{n+1}$ hypotheses of complexity up to $n$.} 
	This, via reasoning as in section \ref{sssec:fundthrm} above, leads to a better generalization guarantee or ``credibility'' of the estimated hypothesis from this class. In this way, Pearl writes in his concluding discussion, ``accepted norms of credibility are correlated with [hypotheses'] simplicity'' (ibid., p.\ 263). However, he immediately adds:
\begin{quote}
From a philosophical viewpoint it is essential to note that in all cases examined the role of \emph{simplicity} was only incidental to the analysis. We would have gotten identical results if instead of $L_c$ being a complexity bounded sublanguage we were to substitute an arbitrary sublanguage with equal number of [hypotheses]. It is not the nature of the [hypotheses] in $I_c$ but their number $|I_c|$ (more precisely, the number of sample dichotomies induced by the members of $I_c$) which affects the various plausibility measures considered.
\end{quote}

In particular, whereas classes of hypotheses of low representational complexity must be small, the converse does not hold. One can select small classes of (representationally) complex hypotheses, and the same capacity-based reasoning for good generalization still applies (cf.\ \citealp[p.\ 65]{Mit97}; \citealp[p.\ 410]{Dom99dmkd}).\footnote{This 
	is the main critique of \citet{Her20pos} of the argument of Blumer et al. Herrmann derives a parallel result for an ``Anti-Occam algorithm'' that selects small-cardinality classes of representationally complex hypotheses.}

	A deeper problem still is that this notion of representational complexity depends on the presupposed formal language and definition of sentence complexity. For any hypothesis that is simple relative to one language, we can design a different language that renders it complex.%\footnote{grue paradox. Priest on gruesome simplicity}
	\footnote{This 
	language-relativity in describing individual hypotheses also clearly arises in some presentations of the \emph{minimum description length} (MDL) approach (e.g., \citealp[sect.\ 6.6]{Mit97}; \citealp[pp.\ 65f]{ShaBen14}). However, these presentations paint a rather, well, simplistic picture of the approach:  in ``refined \textsc{MDL},'' the focus is on the  design of ``universal codes,'' yielding again a robust notion of complexity of \emph{hypothesis classes} \citep{Gru07} that plays a role very similar to capacity notions in statistical learning theory (ibid., sect.\ 17.10).} 
	In that sense representation length does not give us a robust or objective notion of simplicity of individual hypotheses.\footnote{It 
	is sometimes held that ``idealized MDL'' or \emph{Kolmogorov complexity} can offer an objective notion of the representational complexity of individual objects \citep{LiVit08}. See \citet{Ste16pos} for a critique of a suggested justification for Occam's razor via this approach, and \citet[sect.\ 5.2]{Ste18phd} for a critique of its promise of an objective notion of complexity.}
	
Given the bleak prospects for some general mathematical definition to settle what counts as simple for individual hypotheses, one might at this point be inclined to change tack and suggest that in practice, there is often no real problem.	
	 %	In many specific learning problems, some conception of simplicity of hypotheses is already taken for granted and in that sense \emph{given}. 
	 For many specific learning problems, we do appear to have clear intuitions about natural representations or parametrizations of hypotheses. In the standard curve-fitting problem (see, e.g., \citealp[pp.\ 88ff]{Sob15}), where we seek to estimate a polynomial function, there exists a natural parametrization by \emph{degree}.\footnote{Curve-fitting can be cast as a problem in binary classification by treating the curves as hypotheses separating instances with the one label from instances with the other.} The linear functions of degree 1 are simpler than the quadratic functions of degree 2. Moreover, the class of all linear hypotheses is smaller (has lower capacity) than the strict superclass of quadratic hypotheses. Some conception of simplicity of hypotheses is here already taken for granted, which points to a natural  ranking of hypothesis classes, and this ranking neatly aligns with their capacity.\footnote{One
	 might seek to base this conception on some formal definition of simplicity  in terms of number of adjustable parameters, a line going back at least to \citet{Jef39}. But this still does not suddenly give us a robust and objective notion of simplicity of individual hypotheses: the ``grue-like'' problems of representation invariance remain \citep{Pri76pos}. For a recent discussion and critique of defining simplicity by number of parameters, see \citet{Bon23syn}. }
	 
	 If this is a common situation in practice, then, together with the formal connection between low capacity and generalization performance, we may have the basis for an \emph{explanation} of why preferring simple hypotheses generally seems to be a good idea. %\footnote{Pearl 	on explanation}  
	But even if we accept as given, for many specific learning problems, a standard representation or parametrization of hypotheses, the formal connection between low capacity and generalization performance still falls short of constituting a \emph{justification} for preferring simple hypotheses (for these specific learning problems). The issue remains that the theory does not enforce a connection between simple  individual hypotheses (however specified) and classes of low capacity.
	 
%	 ---it can be seen as part of the assumptions we bring to the learning problem. 
%	 
%		 
%	 
%	 
%	 This fits the epistemological perspective developed in sect.\ below. 
%	 
%	 The fact is that we often have such a standard representation or parametrization of hypotheses, that moreover leads to a natural carving out 
%	 
%	 , can give the basis for an 
%	
%	 
%	 Still, even if we were to agree on a standard representation or parametrization of hypotheses, it is a further step to interpret this representation as tracking hypotheses' simplicity, and a further step still to carve out hypothesis classes by these simplicity levels. The fact that we usually do so in practice, in combination with the formal connection between low capacity and generalization performance, could constitute the outlines of an \emph{explanation} of why preferring simple hypotheses seems to be a good idea.\footnote{Pearl 
%	on explanation}  
%	But the formal connection between low capacity and generalization performance falls short of constituting a \emph{justification} for preferring simple hypotheses,  because the mathematics does not enforce a connection between (representationally) simple hypotheses and low capacity.

\subsection{VC dimension as a measure of simplicity}\label{ssec:vccompl}
In contrast to definitions of the complexity of an individual hypothesis, definitions of the capacity of a hypothesis class (like VC dimension) do not depend on a specific representation or parametrization, and do therefore possess a certain objectivity or robustness.\footnote{More
	precisely, the	capacity of a hypothesis class does not depend on how the individual hypotheses are described: all that matters is their data coverage. Language-relativity only turns up when we start redescribing the instance space (cf.\ \citealp[p.\ 482]{Ste09jpl}).}
	But does VC dimension also give an objective or robust measure of the \emph{simplicity} of a hypothesis class? 
	
One might deny this on exactly the grounds that VC dimension does not necessarily align with natural parametrizations of individual hypotheses \citep[p.\ 413]{Dom99dmkd}. In the case of the usual parametrization of polynomials, the higher the number of free parameters, the higher the capacity of the corresponding hypothesis class; but in other cases the two can come apart. The standard example is the class of sine curves $\{ h_\alpha \}_{\alpha \in \mathbb{R}}$ with $h_\alpha(x) = \sin \alpha x$ (ibid.). %; \citealp[pp.\ 72f]{HarKul07}). 
The elements in this class are given by only one parameter (and in that sense the function class is very simple), yet the class has infinite VC dimension \citep[p.\ 78]{Vap00}.\footnote{\citet[p.\ 698]{Vap98} 
	himself writes that since Occam's razor says that the explanation with ``the smallest number of features (free parameters)'' is best, and since this is not supported by theoretical results, ``Occam's razor principle is misleading and perhaps should be discarded in the statistical theory of inference'' (ibid., p.\ 699). Also see \citet[pp.\ 146ff]{CheMul07}.}

Of course, this objection relies on some claim that usual parameterizations do (and exclusively do) track simplicity. %\footnote{An 	 idea that goes back at least to Jeffreys. but critique Ackermann? recent critique by bonk} 	 Such a claim, it seems, will have to based again in some objective definition of simplicity of individual hypotheses, and will run again into the  ultimate non-robustness of representational notions of complexity. But even apart from the formal basis for such a claim, 
But even aside from the ultimate non-robustness of representational notions of complexity, there just exist different and sometimes conflicting intuitions. From one way of looking at it, the class of sine functions \emph{is} maximally complex: exactly because so many possible data configuration can be fit by it (cf.\ \citealp{Rom17pos}). This is the intuition of richness or complexity (or also \emph{falsifiability}\footnote{Vapnik 
	 \citeyearpar[pp.\ 42ff]{Vap00} 	links the capacity of hypothesis classes to Popper's \emph{falsifiability} of theories. Popper famously equated simplicity with falsifiability, and introduced a quantitative notion of falsifiability of theories that he claimed aligned with  number of free parameters. \citet{CorSchVap09jgps} and \citet[p.\ 50ff]{HarKul07} argue that VC dimension is a better measure of falsifiability, though these authors appear to resist linking VC dimension to simplicity. %[in Sch03complexity paper, scholkoepf makes very different connection to complexity
	 })
	made precise in a capacity measure. VC dimension is not \emph{the} measure, but it is \emph{a} plausible and robust measure of the complexity of a hypothesis class. This is enough for my purpose: if a justification is to be had for preferring low capacity, then I think it is reasonable to call this a justification for preferring simplicity---even if there are other reasonable conceptions of simplicity, and even if (to stress again) this notion of simplicity pertains to hypotheses classes and not to individual hypotheses.
%	which does not mean one should forget this is only a particular type of simplicity.

\section{The justification for simplicity}\label{sec:justsimpl}
The fundamental theorem of statistical learning theory ties the simplicity---the VC dimension---of a hypothesis class to its learnability, and indeed already to its learnability by the \textsc{ERM} method. This result offers, first of all, a certain justification for the \textsc{ERM} procedure; although this is a justification with several qualifications, chief among them its \emph{model-relativity} (section \ref{ssec:justerm}). Nevertheless, I will argue that the model-relative justification that learning theory can offer fits right in with a plausible broader epistemological perspective on machine learning methods (section \ref{ssec:epiml}). Finally, I will assemble from all of the previous elements a qualified justification for a simplicity preference (section \ref{ssec:justsimpl}).

\subsection{The justification for empirical risk minimization}\label{ssec:justerm}
%As discussed in section \ref{sssec:fundthrm} above, t
The fundamental theorem shows that \textsc{ERM}$_\mathcal{H}$, for VC class $\mathcal{H}$, is a good method. It is good, and good epistemically, because it satisfies a guarantee of attaining a certain epistemic goal. This guarantee therefore constitutes an epistemic \emph{justification} for the method.

An immediate qualification is that this picture of justification or not---learnability or not---is overly black-white. The quantitative version of the fundamental theorem tells us that a smaller VC dimension leads to a stronger guarantee, making \textsc{ERM} an epistemically better method. So we have a graded guarantee that constitutes a graded notion of epistemic justification. 

But this step from theoretical guarantees to talk about justification comes with several further qualifications still. 

\subsubsection{Qualifications}\label{sssec:justermquals}

%Or that is, so to speak, the opening salvo---there are many things here that need qualification. 

%To start with, the suggestion of a sharp dividing line between justification or no from the fundamental theorem is overly black-and-white. Within the finite VC dimension regime, the size still matters. Smaller VC dimension gives sharper guarantees, and, one could say, a stronger justification for the \textsc{ERM} method. In my discussion I will for convenience mostly adhere to the black-and-white picture, but one should keep in mind that the concept of justification we end up with, like the strength of the bounds of the fundamental theorem, admits of a scale.

A first elementary point is that the fundamental theorem is a mathematical result. Any epistemic justification derived from it, in the context of a real-world learning problem, needs a story how it maps to this learning problem. Most obviously, for any particular real-world problem, a meaningful application of the fundamental theorem (and justificatory claims derived from it) depends on how well the problem can be modelled in the statistical learning theory framework. This includes the match of our prior assumptions with the formal assumption of i.i.d.\ sampling of data from some unknown distribution, but also the match of our goals with the formal choice of the 0/1 error function. The representation of a learning problem in the formal framework of statistical learning theory already forces us to commit to and codify various assumptions \citep[pp.\ 683f]{LuxSch11inc}, and anything that follows from the mathematics should be appraised with an eye to whether these made sense for the original learning problem. 

But even when there are no such modeling concerns, one need not accept that the formal guarantees from the fundamental theorem are sufficiently strong or interesting to warrant talk about justification. There are legitimate reservations one can have about the usefulness of these guarantees.

One possible reservation is that these guarantees are quintessential \emph{frequentist} guarantees. They say something about what we can expect with high probability \emph{before} the sampling and the learning. In that sense we can call methods satisfying these guarantees \emph{reliable} \citep{HarKul07}. But  these guarantees  do not strictly say anything about what we can infer \emph{after} the learning---about the hypothesis that has \emph{actually} been selected, other than that it has been selected by a reliable method \citep[pp.\ 699f]{LuxSch11inc}. For instance, the ``wysiwyg'' guarantee of uniform convergence does not strictly say anything about what we have gotten when we see the result. It is easy to misinterpret such guarantees.
 
A second possible reservation is that the guarantees may be overly  weak. In particular, since the guarantees are agnostic about the true distribution, they are worst-case bounds that for many real-world situations---where we feel we can exclude certain classes of (``pathological'') distributions---would be overly loose or pessimistic \citep[p.\ 680, pp.\ 683f]{LuxSch11inc}.\footnote{That sample sizes can in practice be much smaller was already shown by early experimental results \citep{CohTes92nc}.} 
	This motivates, for instance, the study of ``fast rates'' under further assumptions on the distributions\footnote{Or more precisely, on the relation between the distribution, hypothesis class, and loss function (see \citealp{ErvGruMehReiWil15jmlr}).} and of guarantees that hold for all distributions but are still distribution-dependent.\footnote{This 
	is the idea of the ``theory of universal learning'' of \citet{BHMHY21stoc}. Their motivation is that the usual ``distribution-independent definition of learnability is too pessimistic to explain
practical machine learning,'' showing the need for alternatives that ``better
capture the practice of machine learning, but still give rise to a
canonical mathematical theory of learning rates'' (ibid., p.\ 533).} 
	These studies yield a more complicated picture of  the justification for \textsc{ERM}.
	
Still, %similar to the case of VC dimension as a theoretical notion of simplicity, 
the fundamental theorem gives at least \emph{a} plausible theoretical justification for the method of \textsc{ERM}. This does not exclude that one might (also) wish for different kinds of justification, in any specific problem or in general. In any case, my aim here is to flag the above qualifications, to make clear that accepting the justification for \textsc{ERM}, and indeed the justification for simplicity to follow, presupposes accepting those qualifications. One can reject the argument below simply by rejecting the presuppositions of the statistical learning theory framework. For instance, the reasons \citet{Kel08inc,Kel11incl} offers for rejecting arguments for Occam's razor from statistical learning theory are essentially a rejection of the predictive framework.\footnote{Kelly 
	\citeyearpar[p.\ 329]{Kel08inc} writes that Occam's razor ``should help one to select the true theory from among the alternatives,'' whereas arguments based on risk minimization do not concern ``theoretical truth'' but ``passive prediction'' (ibid., 335). ``But beliefs are for guiding action and actions can alter the world so that the sampling distribution we drew our conclusions from is altered as well'' (ibid.); moreover, ``it is clear that the over-fitting story depends, essentially, upon noise in the data [\dots] One would prefer that the connection between simplicity and theoretical truth not depend essentially upon randomness'' (ibid.). These points are all well-taken (in particular the problem of distribution-shift has recently received more attention, \citealp{WGSARKDC22iclr}), 
but are all already concerns about the scope of statistical learning theory itself.
} 
	But, I will argue, conditional on those qualifications---including the presuppositions of the framework---we can formulate a justificatory argument.%\footnote{With 	an eye to the second reservation, there is nothing here that excludes the possibility of running something very similar to the below  justificatory story for simplicity on the basis of more refined results---I think it would indeed be surprising if this turned out to go awry. But developing this story for the most basic case is enough work for this paper. }

There is, however, a final aspect to this justification that requires more discussion. Namely, it can be seen to cover only one side of an inevitable trade-off.

\subsubsection{The bias-complexity trade-off}
Recall that the notion of learnability of section \ref{sssec:learnty} above concerns learning a hypothesis with \emph{relatively} low risk among those hypotheses in given $\mathcal{H}$. Formally, it concerns finding $\hat{h}$ that minimizes the estimation error $L_\mathcal{D}(\hat{h}) - \min_{h \in \mathcal{H}} L_\mathcal{D}(h)$. 

Intuitively, this concerns the avoidance of \emph{overfitting}. If a hypothesis class $\mathcal{H}$ has overly high capacity, then for any given data sample, the empirically best hypothesis in $\mathcal{H}$ is likely to overfit to random noise in the sample, in which case it is actually significantly worse than the best---lowest-risk---hypothesis in $\mathcal{H}$. Learnability basically concerns the avoidance of such overfitting, and the fundamental theorem then says that overfitting  is avoided if $\mathcal{H}$ is a VC class. 

But this leaves out the other direction of error, namely the \emph{underfitting}. Formally, this concerns the \emph{approximation error}, or the (\emph{absolute}) risk $\min_{h \in \mathcal{H}} L_\mathcal{D}(h)$ of the best hypothesis in $\mathcal{H}$. The absolute risk of the selected hypothesis can be trivially decomposed as the sum of the two types of errors,
\begin{align}
L_\mathcal{D}(\hat{h}) = \underbrace{\min_{h \in \mathcal{H}} L_\mathcal{D}(h)}_{\textrm{approx.\ error}} + \underbrace{L_\mathcal{D}(\hat{h}) -\min_{h \in \mathcal{H}} L_\mathcal{D}(h)}_{\textrm{est.\ error}}.
\end{align}

The opposing pull of these two error terms is also referred to as the \emph{bias-complexity trade-off}. A lower complexity---lower capacity---class excludes more possibilities, and as such embodies, in machine learning terminology, a stronger \emph{inductive bias}.\footnote{This
	is a bit more general than the \emph{bias-variance} trade-off (see \citealp[sects.\ 2.6, 2.9]{HasTibFri09}). Complexity and (inductive) bias are here in the first instance used as informal terms (even if complexity can be made precise as VC dimension, and some authors refer to the approximation error itself as the bias), while the bias and variance in the latter are well-defined statistical terms in regression with mean squared error.} 
	The fundamental theorem yields a guarantee about finding the best in a given class, but this is inevitably a class with some inductive bias. The resulting justification must therefore be relative to this class or inductive bias.\footnote{\label{fn:inbias}Of 
	course, the hypothesis class is not the only---or even the most important \citep[p.\ 684]{LuxSch11inc}---way in which assumptions or biases enter: as discussed in section \ref{ssec:justerm} above, important modeling assumptions must already be made in the formalization of the learning problem (including choice of feature space and loss function). But discussions of inductive bias (in particular around the no-free-lunch theorems introduced shortly) usually assume that these elements are already in place, and center on the further inductive assumptions required.}

\subsubsection{Model-relative v.\ absolute justification}\label{sssec:modelrelvabs}
In the terminology of \citet{SteGru21syn}, this is a \emph{model-relative} justification. It is indeed a justification for a learning method, \textsc{ERM}, that is explicitly  \emph{model-dependent}. The method of \textsc{ERM} is a general procedure---a ``generic learning rule'' \citep[p.\ 68]{ShaBen14}---yet one that  must, by definition, on each application be supplied with further assumptions.  We can view \textsc{ERM} as instantiating a two-place function, that apart from a data sample, also takes a particular hypothesis class. On each specific application, it must be supplied with a hypothesis class or \emph{inductive model} that constitutes further (context-dependent) assumptions, the inductive bias. Correspondingly, the learnability guarantee for \textsc{ERM} is model-relative, because the notion of learnability is relative to a given inductive model or VC hypothesis class. %It is a guarantee of probably finding an hypothesis that is nearly as good as the best hypothesis \emph{in the given class}.

The analysis of Sterkenburg and Gr\"unwald aims to explain how general learning-theoretic guarantees for generic algorithms are consistent with the skeptical import of the so-called no-free-lunch theorems of supervised learning (going back to \citealp{Wol92cs,Wol96nc,Sch94icml}). Modern versions of these results (\citealp[pp.\ 36ff]{ShaBen14}; \citealp[pp.\ 9990f]{SteGru21syn}) say that there can exist no \emph{universal} learning algorithm: every particular algorithm is inadequate in some possible learning situations, situations where \emph{another} algorithm \emph{is} adequate. And since there can be no a priori justification for privileging particular learning situations, so the further interpretation goes, there can be no theoretical justification for any particular algorithm. 

However, % observe that such results presuppose a conception of learning algorithms as purely data-driven or \emph{data-only}: as instantiating a one-place function that only takes data samples. %What the results say is that any data-only algorithm must be inadequate in some situations. %---with the further interpretation that it must therefore lack justification. 
%Thus, 
rather than the generic yet model-dependent \textsc{ERM} algorithm, the no-free-lunch statement applies to \emph{any particular instantiation} of \textsc{ERM} with an inductive model $\mathcal{H}$, any particular one-place ``data-only'' function \textsc{ERM}$_\mathcal{H}$. The no-free-lunch result of \citet[thrm.\ 5.1]{ShaBen14} essentially states that for any specific inductive model $\mathcal{H}$, the data-only algorithm \textsc{ERM}$_\mathcal{H}$ is inadequate (i.e., with high probability suffers high absolute error) for some situations (i.e., for some true distributions; informally, those that do not match $\mathcal{H}$'s inductive bias) where another data-only algorithm, like \textsc{ERM}$_\mathcal{H'}$ for \emph{another} inductive model $\mathcal{H}'$ (that \emph{does} match the situation), is adequate.\footnote{Another
	way of casting this result is that the class of \emph{all} classifiers is not learnable: since, for any possible distribution, this class has minimum approximation error, its learnability (guarantee of low estimation error) would guarantee low absolute error for any possible distribution. In fact, the proof of \citet[thrm.\ 5.1]{ShaBen14} already shows the failure of learnability of classes with infinite VC dimension, and the no-free-lunch theorem is in their presentation part of the proof of the fundamental theorem (ibid., pp.\ 45ff).}

In other words, while we can have a model-relative justification for model-dependent algorithms (of the rough form, ``for any instantiated inductive model of the right form, works well relative to the model''), there is no \emph{absolute} justification (``works well whenever'') for any inductive model. The impossibility of such absolute justification is also an important part of the argument of \cite{Dom98kdd,Dom99dmkd} against a possible justification of Occam's razor.

\subsubsection{The failure of an absolute justification for simplicity}\label{sssec:noabsjust}
Domingos takes issue with what he calls the ``second razor'': that ``given two [hypotheses] with the same [empirical] error, the simpler one should be preferred because it is likely to have lower generalization error'' \citeyearpar[p.\ 410]{Dom99dmkd}. (To be distinguished from the ``first razor,'' that ``the simpler [hypothesis] should be preferred because simplicity is desirable in itself,'' ibid.)\footnote{We
	are here, of course, likewise concerned with an \emph{epistemic} justification for Occam's razor, not with claims that simplicity is better because it points to hypotheses or theories that are easier to work with or more aesthetically pleasing (cf.\ \citealp[pp.\ 58f]{Sob15}).} 
	He writes that theoretical ``zero-sum arguments''---no-free-lunch theorems---``imply that the second razor cannot be true'' \citeyearpar[p.\ 413]{Dom99dmkd}. Namely, ``they imply that, for every domain where a simpler [hypothesis] is more accurate than a more complex one, there exists a domain where the reverse is true, and thus no argument which is preferable in general can be made'' (ibid.). 

This is true, and an expression of the impossibility of absolute justification, applied to the choice of a simple single hypothesis.\footnote{This 
	point has also been brought out experimentally (e.g., \citealp{Sch93ml,Web96jair}).} 
	However,  I already observed that the relevant notion of simplicity attaches to hypothesis classes, not single hypotheses. 
Domingos's critique of the theoretical ``PAC-learning argument'' for the second razor, which is also a no-free-lunch observation, is more relevant: 

\begin{quotation}
``[uniform convergence results]\ only say that if we select a sufficiently small set of [hypotheses] prior to looking at the data, and by good fortune one of those [hypotheses] closely agrees with the data, we can be confident that it will also do well on future data. The theoretical results give no guidance as to how to select that [hypothesis class].''
\citeyearpar[p.\ 410]{Dom99dmkd}
\end{quotation} 

This is again true, and an expression of the impossibility of absolute justification, applied to the choice of (\textsc{ERM} instantiated with) a simple hypothesis class.  For a sufficiently small (low-capacity) set of hypotheses, the fundamental theorem gives a guarantee of probably finding the near-best hypothesis in the class. But this hypothesis is only good in an absolute sense (has low risk and so ``will also do well on future data'') if the class of hypotheses was good to begin with (contains a hypothesis that has low risk and so ``closely agrees with the data''). By the no-free-lunch results, we know that for any hypothesis class there are learning situations such that the class is not good. And the theory does not guide us towards a good hypothesis class prior to looking at the data.

In sum, an absolute justification of (\textsc{ERM} with specific) choice of simple hypothesis class is impossible. Nevertheless, this leaves a model-relative justification of \textsc{ERM}. Such a justification, I will now argue, is still of epistemological interest, and indeed points towards a qualified justification of simplicity.  

\subsection{Epistemology and machine learning theory}\label{ssec:epiml}

\subsubsection{``Model-relative'' epistemology}\label{sssec:pedigreevsforward}
Absolute justification is central to a general epistemological project where we are concerned with the foundations of our knowledge. This is a project of turning back, of retracing the justificatory basis for a statement or belief  of interest. In the context of machine learning algorithms, we ask: what is the basis, the justification, for trusting what our learning algorithm returns? By the no-free-lunch theorems, we know that our learning algorithm's outputs are grounded in a particular inductive bias. So we are led to ask: what is the foundation, the justification, for this inductive bias? Accepting the Humean argument that neither deductive nor nondeductive justification is forthcoming, we are ultimately led to skepticism (\citealp[pp.\ 9992ff]{SteGru21syn}).

Much of modern philosophy of science implicitly or explicitly views this general epistemological project as a dead end. %A broad reaction perceivable in the philosophy of science to the skeptical problems of induction and underdetermination is a dismissal of this general epistemological project. %Several philosophers of science have proclaimed this general epistemological project a dead end. %taken positions that amount to a dismissal of this general epistemological project. %Peirce's model of inquiry rejects the Cartesian demand for justification for beliefs that are not in actual fact subject to ``real and living doubt,'' 
Reichenbach, in the words of \citet[p.\ 254]{Fra00ppr}, thought that empiricist epistemology must reject ``Rationalism's stringent criterion of adequacy: that an epistemology must show how absolutely reliable knowledge is possible.'' Van Fraassen \citeyearpar{Fra89,Fra00ppr,Fra04} himself offers an outlook of an empiricism  in explicit rejection of ``defensive epistemology'' which ``concentrates on justification, warrant for, and defence of one's belief'' \citeyearpar[p.\ 170]{Fra89}.
Peirce, in the words of \citet[p.\ 177]{Lev98inc}, ``explicitly dismissed doxastic skepticism when he observed that merely writing down a question challenging some current assumption is not sufficient to create the sort of doubt that should occassion an inquiry.'' Levi \citeyearpar{Lev80,Lev04} himself further develops Peirce's doubt-belief model of inquiry in explicit rejection of ``pedigree epistemology,'' under which ``one is obliged to justify current beliefs'' \citeyearpar[p.\ 11]{Lev04}.   
%thought that empiricist epistemology must shake off the ``fundamental thesis of rationalism, according to which genuineknowledge has to be as reliable as mathematical knowledge'', and van Fraassen 
% Levi speaks of \emph{pedigree epistemology} and van Fraassen of \emph{defensive epistemology} (\citeyear[p.\ 170]{Fra89}; \citeyear[p.\ 371]{Fra07inc}); and both offer alternative outlooks. 
The broad alternative project that arises is a pragmatist one where we take seriously that one will always already start with a body of beliefs that one at that time does not actively or genuinely doubt. The interesting question is not whether one actually has an ultimate justification for these beliefs. The interesting question is how to proceed from these beliefs: how to \emph{improve} these beliefs. This leads to an epistemological project that investigates how to improve (refine, revise, update) beliefs in the light of new data. And in this project there is still an interesting question of justification: we can theorize about better and worse ways of doing so.\footnote{This
	broad epistemological approach is certainly not particular to the authors I sampled here. The repudiation of quests for absolute justification is found in the writings of many other prominent philosophers, like Popper (``[t]he piles are driven down from above into the swamp,'' \citealp[p.\ 94]{Pop59}); the ``model-relativity'' of all our knowledge is not just central to modern authors on induction like Howson \citeyearpar{How00}, but already to Carnap and indeed to Kant.}\textsuperscript{,}\footnote{\citet[pp.\ 86f]{Sob15}
	also sets apart, with reference to Neurath's boat, a ``foundationalist'' picture from a ``more defensible'' alternative picture where we ``now have numerous beliefs about the world. Our task is to take new observations into account so as to improve our system of beliefs. We don't start from zero; we start from where we are'' (ibid., p.\ 87). Notably, though, he presents these two pictures as two different versions of \emph{Bayesianism}; there is no reference to these different epistemological views in his discussion of frequentist ideas. Other authors, including Levi and van Fraassen, also assume a Bayesian framework. %[Mayo similar philosophical perspective from frequentist ideas; mention of Salmon] 
I do not do that here: I am not after a formal account of the wider epistemic context of agents using machine learning methods to inform their beliefs. I am here interested in the epistemological lessons we may draw from a formal account of the machine learning methods themselves, namely, machine learning theory.}

The same sentiment can be found in the machine learning literature. Already thirty years ago, \citet[p.\ 45]{Rus91ps}, after discussing the infeasibility of ``tabula rasa'' learning, writes that ``the picture that is currently fashionable in machine learning is that of an agent that \emph{already knows something} and is trying to learn some more.'' In similar vein, \citet[p.\ 81]{Dom12cacm} himself, after attributing to Hume and Wolpert the insight that ``data alone is not enough,'' writes that ``induction (what learners do) is a knowledge lever: it turns a small amount of input knowledge into a large amount of output knowledge.'' Importantly, these authors, like also \citet[p.\ 94]{ShaBen14}, do not just point out the impossibility of inductive inference without assumptions. They presuppose that there is always a starting point of initial knowledge, and put to one side the question of the actual basis for this supposed knowledge. They instead take machine learning to be about how to best proceed from initial knowledge: how to learn more, or turn input knowledge into more output knowledge. %In this picture, machine learning is about how to design methods that improve starting assumptions with new data. 

The model-relative guarantees derived within the theory of machine learning serve exactly such a perspective. Model-relative guarantees concern algorithms that presuppose the instantiation, in each application, of an inductive model, that codifies specific prior knowledge. And these guarantees show that such algorithms are good relative to the instantiated inductive model, relative to the prior knowledge. This fits a picture where any real-world learning problem arises in a context where we already take many things for granted, and are willing to accept as prior knowledge. Given our starting point (our particular learning problem and goal, and the way we codify prior knowledge in a formal inductive model), the relevant theoretical model-relative guarantee advises us on how to proceed (what model-dependent algorithm to use). Learning-theoretic results thus provide a normative component to this general 
epistemological perspective on learning methods.\footnote{Following
	up on footnote \ref{fn:inbias}, prior knowledge is not only instantiated in the inductive model qua hypothesis class, but also already in the other formal components of the learning problem. But clearly learning-theoretic guarantees must always be relative to these choices as well.}

\subsubsection{The theory and the practice}
In evoking a wider epistemological perspective on machine learning, we have a responsibility to do justice, not only to the normative role of the theory, but also to the actual practice of (modern) machine learning. Here arises a worry that the preceding picture is excessively neat. For one thing, it is not exactly standard machine learning practice to carefully specify an inductive model strictly on the basis of well-formulated domain-specific assumptions; the practice, say in deep learning, is to a significant extent one of trial-and-error of different general architectures and hyperparameters, that codify largely ill-understood inductive biases. For another, the kind of justification that practitioners offer is less based on the (formal) properties of the learning algorithm than on the empirical performance of the output classifier (the trained model), in the first instance on a separate test data set or in a cross-validation procedure.\footnote{This
	turn away from explicit modeling and towards predictive performance is what is indeed often seen to separate machine learning from ``traditional'' statistical inference \citep{Bre01ss}.}
	
A full reply to this worry would have to engage more systematically with the relation between the theory and the practice of machine learning, which I do not attempt in this paper. Here I will just observe the following. Whatever the several back-and-forths of design and evaluation in an actual machine learning pipeline, the core remains the training of an algorithm to return a classifier that generalizes well. And if there is one theoretical lesson that will always stand, then it is that the algorithm must possess restrictive inductive biases, and that the algorithm will only do well if the inductive biases are appropriate.
It is also nothing short of a practical necessity to constrain, amidst all the further guesswork and trial-and-error, the possible inductive models to at least some extent; and this will still always involve at least some amount of knowledge about what is likely to be appropriate in the current problem. Going further, we can make the minimal normative point that it is indeed \emph{better} to try and implement inductive biases that are aligned with what we believe or are prepared to assume about the relevant domain. With all the qualifications and idealizations involved in linking the theory to the practice (which also ties back to the qualifications listed in section \ref{sssec:justermquals} above), if there is a normative role for the theory to play, then it is in grounding learning procedures that can capitalize on an inductive model encoding inductive assumptions, fitting in a broad ``model-relative'' epistemological picture on machine learning methods.

\subsubsection{``Means-ends'' epistemology}\label{sssec:mee}
The idea that theoretical learning guarantees provide the basis for a normative epistemology is also central to the philosophical tradition best known as \emph{formal learning theory} \citep{Kel96,Kel16inc,Sch17sep,Gen18phd}. I will briefly make a connection to this tradition, in order to identify another crucial ingredient for the qualified justification for a simplicity preference that I will propose after.

%The preceding account of model-relative justification within a forwards-looking perspective bears much affinity to the philosophical approach best known as \emph{formal learning theory} \citep{Kel96,Kel16inc,Sch17sep}. 
The core principle of formal learning theory, which also has its roots in machine learning theory,\footnote{Specifically,
	the approach of \emph{algorithmic learning theory} going back to \citet{Put65jsl} and \citet{Gol67ic}; see \citet{JaiOshRoySha99}.} 
	is that inductive problems call for a context-dependent \emph{means-ends} analysis of what epistemic notions of success (ends) are attainable with what assumptions and methods (means). %\footnote{alternative term, logical reliability. differences}
 Schulte \citeyearpar{Sch99bjps} therefore also speaks of ``means-ends'' epistemology. 

This means-ends analysis is context-dependent in the sense that given a particular learning problem, which usually comes with restrictive ``background assumptions,'' the analysis does not question these assumptions (\citealp[p.\ 11]{Kel96}; \citealp[sects.\ 1.3, 2.2]{Sch17sep}). For a particular learning problem and notion of success, the analysis is concerned with showing that certain methods can or cannot ``solve'' the problem (can or cannot attain the notion of success), given the background assumptions. This again fits a ``model-relative'' epistemological perspective (cf.\ \citealp[pp.\ 713f]{Kel16inc}), where the analysis provides a model-relative justification (with the inductive model constituted by the background assumptions) for methods that solve the problem.

However, this ``problem solvability analysis'' of whether and which methods can solve a given learning problem is not the only possible direction of theoretical analysis \citep[p.\ 37f]{Kel96}. Indeed, each of the different parameters at play (the learning problem, the background assumptions, the notion of success, the methods) we can either vary or keep fixed  \citep[p.\ 696]{Kel16inc}. In particular, we can fix a learning problem and notion of success, and ask what background assumptions are needed for a method to possibly solve the problem. Here we are after characterization results that give necessary and sufficient conditions for the attainability of (i.e., the existence of a method that attains) the relevant notion of success \citep[p.\ 74]{Kel96}.  In the words of Kelly (ibid.), 
\begin{quote}
a characterization theorem isolates exactly the kind of background knowledge
necessary and sufficient for scientific reliability, given [\dots]\ %the interpretation of the hypotheses and 
the sense of success demanded. To revive Kant's expression,
such results may be thought of as \emph{transcendental deductions} for reliable inductive
inference, since they show what sort of knowledge is necessary if reliable
inductive inference is to be possible.
\end{quote}

In the logical framework studied in formal learning theory, there arises a neat hierarchy where different notions of success are characterized by the topological structure of the problem and background assumptions \citep{Kel96}. In the statistical learning theory framework, the fundamental theorem characterizes the main notion of success in terms of the combinatorial structure of the background assumptions: for a method to have the success guarantee of learnability, the hypothesis class must have finite VC dimension. Thus, adopting Kelly's words, the fundamental theorem shows what sort of knowledge (form of hypothesis class) is necessary for reliable inductive inference (a learnability guarantee). This provides a means-ends reason for modelling, if we can, our background assumptions in the form of a class of hypotheses with finite VC dimension, a simple class of hypotheses. From the combination of a model-relative perspective and a means-ends analysis thus arises a justification for preferring simplicity---albeit with important qualifications.

%Thus, when we face a certain formal learning problem, and we would like a certain guarantee of learning success, the relevant characterization result tells us which form the formal background assumptions must take. This ``transcendental deduction'' gives us a means-ends justification for modeling, if we can, the informal knowledge we bring to the formal learning problem in the shape of formal background assumptions of the requisite form. 

%Note that this notion of ``reliable inductive inference'' is still model-relative: it is relative to the knowledge or background assumptions.  

%``learning theory is not in the business of questioning assumptions'' (p.\ 11)

\subsection{A qualified justification for simplicity}\label{sssec:qualjustunif}\label{ssec:justsimpl}
%The general line of reasoning that arises from the previous discussion goes as follows. 
%
%We face a certain learning problem for which we want to use a formal learning method or machine learning algorithm.  In line with the forwards-looking perspective, we come to this problem already equipped with (informal) prior knowledge; and we would like the learning method we choose to have a certain guarantee of success, relative to our prior knowledge. %Moreover, we would like this learning method to have a certain guarantee of success. and we are interested in a success guarantee relative to our prior knowledge. 
%More precisely, we would like to use a learning method that has a guarantee relative to a formal inductive model that matches our prior knowledge. Now the relevant learning-theoretic characterization result for this type of guarantee tells us which shape the formal background assumptions must take. Only when the formal background assumptions satisfy this condition, does there exist a method we can use that has the desired model-relative guarantee. This ``transcendental deduction'' gives us a means-ends justification for modeling, if we can, our prior knowledge in the shape of formal background assumptions of the requisite form.

%The specific reasoning in the case of simplicity and uniform convergence in SLT then goes as follows.

We can now assemble from the building blocks of the previous sections a justificatory argument.

\subsubsection{The argument}
 We face a certain problem of classification, which we are prepared to model as a problem in statistical learning. We enter this problem with further prior knowledge still; and we are interested in a method that is good relative to this prior knowledge. As a formalization of what it means for a method to be good relative to prior knowledge, we adopt the model-relative notion of learnability. Now the fundamental theorem tells us that for there to exist a method with this guarantee of learnability, we need to formulate a hypothesis class, as formalization of our prior knowledge, that is a VC class---that is \emph{simple}. Only when the hypothesis class is simple, does there exist a method with the guarantee of learnability relative to this hypothesis class. This ``transcendental deduction'' gives us a means-ends justification for modeling, if we can, our prior knowledge in the shape of a simple class of hypotheses.%\footnote{more 	ambitious [and more properly transcendental] argument: ML methods work well in absolute sense [manifest feature of experience], so justification for simple models that they start with. connection evolutionary arguments in e.g. duda et al...?} %\texttt{smth more about with a simple class we have a better estimate of the error of selected predictor; also in that sense more reliable} 
 
The previous reasoning was based on the black-and-white picture of learnability or no. The quantitative version of the fundamental theorem offers a more fine-grained version; but the essence of the argument is the same. Accepting the guarantee of learnability, we recognize that stronger bounds give a stronger guarantee, and we take a method with a stronger guarantee to be better. Now the quantitative version of the fundamental theorem tells us that a VC class of lower VC dimension---a \emph{simpler} hypothesis class---gives a stronger guarantee. This gives us a means-ends justification for modeling, to the extent we can, our prior knowledge in the shape of class that is maximally simple (of maximally low VC dimension).

\subsubsection{Qualifications}
The above argument comes with a series of presuppositions and restrictions in scope, most of which I have discussed earlier. The argument only applies to learning problems that fit the statistical learning theory framework, and it presumes that a learning method is (more) justified if it satisifies (stronger bounds for) the formal criterion of learnability (section \ref{ssec:justerm}). In particular, it presumes the epistemological value of model-relative justification (as I have argued for in section \ref{ssec:epiml}). And it presumes that the VC dimension of a hypothesis class is a plausible criterion of simplicity (as I have argued for in section \ref{ssec:vccompl}).

But perhaps most importantly, the final step of the argument still comes with a crucial qualifier. We have a means-ends justification for modeling our prior knowledge in the form of a simpler class, \emph{to the extent we can}. The theory can be seen to push into one direction, towards simplicity. However, %there is also a pull from the opposite direction. T
the actual context of the learning problem, and in particular our prior knowledge, acts as a check on this push towards simplicity.

%generally pulls away from simplicity, in the direction of weaker assumptions, of a more complex hypothesis class. 

%But perhaps most importantly, the final step of the argument still comes with a crucial qualifier. We have a means-ends justification for modeling our prior knowledge in the form of a simple class, \emph{if we can}. This leaves the cases where we plausibly cannot. These will be cases where our prior knowledge is too weak: where we cannot plausibly restrict the classifiers that we think could be accurate to a class of finite VC dimension.

The prior knowledge we have in each instance is a crucial constraint in the argument. In line with the epistemological ``model-relative''  perspective sketched above, we always enter a learning problem with informal prior knowledge (beliefs we have, assumptions we are willing to make), and it is this informal prior knowledge that we need to codify in some formal hypothesis class. There will always be some leeway here, simply because it will be a significant translation step from our informal knowledge to the formal object of a hypothesis class, but it is still a constraint. It will never be  reasonable, for instance, to adopt a singleton hypothesis class (a maximally simple class of VC dimension 0), because that would mean there would not be a learning problem to begin with. Without this constraint by informal prior knowledge, the argument would lose the connection to the epistemic context of an actual learning problem, and collapse into the useless advice to always choose a class of VC dimension 0. 

The downside is that there will be learning problems where this constraint, this check from complexity, is too strong for the simplicity argument to still be meaningful. There will be situations, in particular, where our prior knowledge is too weak to plausibly translate  into a class of VC dimension sufficiently small to still yield useful bounds. If such situations are the rule rather than the exception, then this seems to be a serious restriction in the scope of the argument.

The good news is that we can embed the argument into a more general setting, presupposing a much weaker kind of prior knowledge. Namely, in rough terms, we can simultaneously evaluate a (ranked) sequence of \emph{multiple} VC classes, where the theoretical push towards simplicity manifests itself again in a preference for lower-VC-dimension classes; but one which is now checked by the classes' empirical error. % certain trade-off between VC dimension and empirical error. In fact, this trade-off 
In fact, this gives a trade-off which can be automated into a learning procedure. This is Vapnik's second ``inductive principle,'' or the method of \emph{structural risk minimization}.

\subsubsection{Structural risk minimization}
Indeed, rather than on uniform convergence and \textsc{ERM}, discussions of Occam's razor in statistical learning theory  tend to focus on this method (\citealp[ch.\ 3]{HarKul07}; \citealp[p.\ 335]{Kel08inc}; \citealp[sect.\ 7.3]{ShaBen14}; \citealp{BarCevGne22mam}). Moreover, this method is directly applicable to the problem of model selection, which takes center stage in the modern philosophical debate about Occam's razor \citep{ForSob94bjps,Sob15}. Very briefly, the formal route to the method of structural risk minimization (\textsc{SRM})  is the following. 

We start with a generalization of uniform convergence (definition \ref{def:uniconv} above). This generalization applies to a weighted countable sequence of VC classes, and gives a uniform accuracy bound for each hypothesis which depends on the class that the hypothesis is in \citep[thrm.\ 7.4]{ShaBen14}. Next, similar to how the \textsc{ERM} method is defined as minimizing a uniform convergence bound, which leads to a minimization of empirical risk (section \ref{sssec:unifconv} above), the \textsc{SRM} method is defined by minimizing a generalized uniform convergence bound, which leads to a minimization of a function of empirical risk \emph{and} VC dimension (ibid., p.\ 62). This, finally, translates into a bound on \textsc{SRM}'s performance; and we can also prove a certain weaker guarantee of \emph{nonuniform} learnability (ibid., def.\ 7.1, thrm.\ 7.5).

I think that \textsc{SRM} can be understood as implementing a parallel application, over multiple VC classes at the same time, of the core argument for simplicity that I gave above. The theoretical push towards simplicity is again a push towards hypotheses from classes of lower VC dimension. This push is again checked by the classes' adequacy for the learning problem, or the adequacy of the inductive assumptions they codify. However, this is now not done by an informal evaluation of how well the formal inductive assumptions match our background knowledge. Instead, this adequacy is directly estimated by empirical error. This gives rise to a quantitative trade-off between these two elements, which \textsc{SRM} automates. By taking into account empirical errors, \textsc{SRM} can thus be seen to automate the evaluation of the adequacy of the inductive assumptions in individual VC classes; but of course the whole procedure is still relative to an initial choice of a weighted sequence of VC classes, which now constitutes the inductive model.

Clearly, it needs more work to spell out this view in the proper amount of detail, including how it fits with earlier discussions of Occam's razor in \textsc{SRM}, and with the philosophical discussion of Occam's razor in model selection methods. 
But that work must wait for another occasion.%\footnote{\label{fn:modelassess2}As 	mentioned in footnote \ref{fn:modelassess1}, in my argument I focused on the epistemic end of learnability, pertaining to the training or model fitting stage. This is the earlier stage in a machine learning procedure, typically followed implicitly (by noting training error) or explicitly (by estimating performance with a test set) by model assessment, and possibly still an intermediate model selection (\citealp[ch.\ 7]{HasTibFri09}; \citealp[ch.\ 14]{ShaBen14}). The core argument, based on \textsc{ERM}, can thus be seen to be restricted to the earlier step; but a

\section{Conclusion}\label{sec:concl}
In this paper, I described a means-ends %model-relative 
justificatory argument for Occam's razor from statistical learning theory, based on the method of empirical risk minimization (\textsc{ERM}). I think this argument accomodates both intuitions and arguments in the machine learning literature in favor of the possibility of such a justification (by actually providing one) and those against (in its various qualifications). It is an honest epistemic justification, that connects a simplicity preference to guarantees of predictive accuracy; and it does not rely on any extra-theoretical assumptions about the true or best hypotheses being simple. But it does presuppose acceptance of the justificatory force of the statistical learning theory framework, including the model-relative nature of the theoretical guarantees. And both the notion of simplicity (as pertaining to classes rather than individual hypotheses) and the means-ends nature of the argument (pushing for simplicity as a constraint on inductive assumptions, rather than directly on inductive conclusions) %(in one line, to prefer simpler models in order to obtain better model-relative guarantees) 
is perhaps different and more minimal than what one might have hoped from an argument for Occam's razor. %But this, I hope to have made clear, is what it is.

%and this model-relativity, I have argued, aligns with the adoption of a broader epistemological perspective that moves beyond the hope of absolute justification. %Furthermore, the arguably seriously limited scope of the ERM method does translate in an equally limited scope of the argument.
%but this, I have argued, is what it is.

In its essence, the argument capitalizes on a push from the mathematical theory towards specifying a simple hypothesis class. The theoretical push towards simplicity, however, is checked or indeed opposed by the informal knowledge one brings to the learning problem, which will generally rather pull in the direction of complexity. This points at a certain trade-off, which is in fact automated in 
%This means that in those situations in which one's assumptions do not plausibly allow for a hypothesis class that is sufficiently simple (that still has meaningful theoretical guarantees), the theoretical push towards simplicity, and so the argument, appears moot. However, the reasoning and theoretical results underlying \textsc{ERM} form the core of a more general approach, accommodating weaker assumptions. This approach is automated in 
the method of structural risk minimization (\textsc{SRM}); and this methods thus appears to directly implement a simplicity preference in inductive inference. It is indeed this method that is usually brought up in discussions  of Occam's razor in statistical learning theory, also as representative for a wider family of regularization techniques in machine learning; further, it is closely related to the statistical methods for model selection discussed in the modern philosophical debate. To complete the case that the current account accomodates earlier arguments and intuitions pro and con, it therefore needs to be spelled out how exactly it figures in the method of \textsc{SRM}. This I intend to do in future work; but I think the core argument is already here.

\begin{comment}

\begin{align*}
x^2 + y^2 &= 1
\\ y &= \sqrt{1 - x^2}.
\end{align*}

\end{comment}

\small

\DeclareRobustCommand{\VAN}[3]{#3} % proper Dutch 'van/de' capitalisation

%\bibliographystyle{abbrnvatnoaddress}
%\bibliography{all}

\end{document}